# How Effectively Can BERT Models Interpret Context and Detect Bengali Communal Violent Text?

Abdullah Khondoker[a], Enam Ahmed Taufik[a], Md. Iftekhar Islam Tashik[a], S M Ishtiak Mahmud[a] and Farig Sadeque[a]

[a]*Department of Computer Science and Engineering, BRAC University, Kha 224 Bir Uttam Rafiqul Islam Avenue, Merul Badda, 1212 Dhaka, Bangladesh*

ABSTRACT

The spread of cyber hatred has led to communal violence, fueling aggression and conflicts between various religious, ethnic, and social groups, posing a significant threat to social harmony. Despite its critical importance, the classification of communal violent text remains an underexplored area in existing research. This study aims to enhance the accuracy of detecting text that incites communal violence, focusing specifically on Bengali textual data sourced from social media platforms. We introduce a fine-tuned BanglaBERT model tailored for this task, achieving a macro F1 score of 0.60. To address the issue of data imbalance, our dataset was expanded by adding 1,794 instances, which facilitated the development and evaluation of a fine-tuned ensemble model. This ensemble model demonstrated an improved performance, achieving a macro F1 score of 0.63, thus highlighting its effectiveness in this domain.

In addition to quantitative performance metrics, qualitative analysis revealed instances where the models struggled with context understanding, leading to occasional misclassifications, even when predictions were made with high confidence. Through analyzing the cosine similarity between words, we identified certain limitations in the pre-trained BanglaBERT models, particularly in their ability to distinguish between closely related communal and non-communal terms. To further interpret the model's decisions, we applied LIME (Local Interpretable Model-Agnostic Explanations), which helped to uncover specific areas where the model struggled in understanding context, contributing to errors in classification. These findings highlight the promise of NLP and interpretability tools in reducing online communal violence. Our work contributes to the growing body of research in communal violence detection and offers a foundation for future studies aiming to refine these techniques for better accuracy and societal impact.

## 1. Introduction

Communal violence refers to conflicts between different ethnic or communal groups, often fueled by group loyalty and inter-group animosity. Such violence is characterised by targeted attacks based on group identity. It poses a significant threat to social harmony and stability, particularly in countries like Bangladesh, where the population is both diverse and densely packed. The widespread adoption of digital media and communication platforms has intensified this problem, with over 44.7 million people in Bangladesh actively engaging in social media [20]. This increase in online activity has facilitated the spread of hateful and inflammatory content, commonly referred to as cyber hate. Such content has already triggered numerous instances of physical violence against various ethnic, religious, and social communities, leading to widespread social unrest.

<Corresponding author.
✉abdullah.khondoker@g.bracu.ac.bd (A. Khondoker);
enam.ahmed.taufik@g.bracu.ac.bd (E.A. Taufik);
iftekhar.islam.tashik@g.bracu.ac.bd (Md.I.I. Tashik);
sm.ishtiak.mahmud@g.bracu.ac.bd (S.M.I. Mahmud);
farig.sadeque@bracu.ac.bd (F. Sadeque)
🌐 https://www.abdullahkhondoker.info/ (A. Khondoker);
https://sites.google.com/view/enamahmedtaufik (E.A. Taufik);
https://www.bracu.ac.bd/about/people/farig-yousuf-sadeque (F. Sadeque)
ORCID(s): 0009-0007-4665-5298 (A. Khondoker); 0000-0001-6797-7826 (F. Sadeque)

In South Asia, the influence of social media has contributed to frequent incidents of communal violence. Countries in this region are often characterised by religious divisions, with populations that hold strong religious sentiments [30]. Bangladesh, a country where diverse religious communities coexist, is predominantly Muslim, with Hindus and other religious minorities making up the remainder. These minorities often harbour fears of persecution by the majority, leading to societal trust issues. For example, in 2021, during the Hindu festival of Durga Puja, a significant event for Hindus, an image circulated on social media purportedly showing the holy book of Islam, the Quran, placed in a Hindu temple. The image quickly went viral, inciting violence between Muslim and Hindu communities. The violence resulted in at least 150 injuries and attacks on 80 temples across the country [4].

The events in Bangladesh illustrate the vulnerability of these societies to communal violence and the role of social media in escalating tensions, often without substantiated evidence. Although considerable research exists on the role of social media in spreading hate speech and influencing social issues, there remains a critical gap in developing specialised models that can detect and analyse comments specifically inciting communal violence. Addressing this gap is essential to understanding and mitigating the role of online discourse in triggering real-world violence.

This research aims to address this gap by focusing on the critical task of identifying text that incites communal violence. The primary research question guiding this study is whether state-of-the-art Natural Language Processing (NLP) tools can accurately classify and target communal violence, given the complexities of language and context. Classifying such violent text is a challenging task, especially in the Bengali language, which is rich in semantic nuance and often features sarcasm, making harmful content difficult to detect.

To address this question, we propose the following hypotheses:

1. Fine-tuning BanglaBERT may improve its performance in classifying communal violence, as it has shown effectiveness in detecting Bangla hate speech and cyber hate.
2. Comments categorized as Noncommunal cannot simultaneously belong to another class, although comments in other classes may overlap.
3. Our dataset is multi-class, which will aid our model in understanding and identifying different types of violent text more accurately.
4. In real-world scenarios, comments may be ambiguous or require context to be properly understood. In such cases, our proposed model may risk misclassification.

To achieve our objectives, we will utilise a comprehensive dataset containing various types of textual content, such as comments and posts collected from social media platforms. Experts have categorised this data to cover a wide range of communal violent texts and topics relevant to this context. This categorization will aid in building robust NLP models and facilitate the analysis. We will employ state-of-the-art machine learning techniques and advancements in NLP to train these models on our extensive dataset. The models will be designed to detect communal violent text, capturing subtle linguistic nuances and contextual clues.

Our ultimate goals are twofold: first, to determine whether a given text can be categorised as communal violent text, and second, to identify the specific category within the spectrum of communal violent text. We expect that achieving these objectives will not only fill the current research gap but also support proactive measures to prevent communal violence by deepening our understanding of the linguistic cues that fuel such conflicts.

## 2. Literature review

### 2.1. Contextual relevance

Communal violence is defined as violent conflicts between different ethnic, religious, or communal groups within a society. The underlying motivations frequently include a desire to dominate another community, exact revenge for perceived injustices, or gain political power. Several studies have explored the impact of communal violence on society and compared deaths resulting from communal violence to those caused by civil wars [11].

The risk of communal violence is especially high in South Asia, a region with a diverse ethnic and religious population living close together. In recent years, such violence has occurred in several parts of the region, including India, Myanmar, and Bangladesh. Communal conflict has been an ongoing concern in India, with tensions between various religious, caste, and class groups persisting since independence. Political strategies frequently exploit these divisions, as evidenced by movements such as 'non-Brahminical Hindutva', which seek to unite lower castes under a broader Hindu identity, fostering prejudices against Muslims. The presence or absence of such movements, combined with historical caste divisions, has a significant impact on whether communal violence is mitigated or fueled, as seen in the contrasts between Uttar Pradesh and Gujarat. Wilkinson also investigates how political competition and backward caste mobilisation, particularly in southern states, affect the dynamics of communal conflict [30].

Similar dynamics are evident beyond India, particularly in Rakhine State, Myanmar, where the ethno-religious conflict between the Buddhist majority and the Muslim minority Rohingya has escalated since 2016, with severe consequences for the political landscape [28].

Bangladesh has also experienced various forms of communal violence. While Muslim-Hindu tensions have been a significant source of conflict, particularly during incidents like the post-election violence in 2001 [25] and the Durga Puja attacks in 2021 [4], the issue extends beyond these two communities. The Rohingya crisis, which forced over a million Rohingya Muslims to flee persecution in Myanmar and seek refuge in Bangladesh, has added another layer of tension, especially in the already resource-strained regions of Cox's Bazar. Additionally, the Chittagong Hill Tracts (CHT) region has been a focal point of violence, particularly between the aboriginal communities and the majority of Bengali settlers. In February 2010, violence between these groups resulted in the burning of 357 houses across 11 villages, leaving thousands of ethnic minorities homeless [8].

In the digital age, social media has become increasingly influential in shaping public perceptions and actions. In South Asia, several recent violent incidents have been exacerbated by social media, which played a crucial role in intensifying these events and contributing to communal violence [31, 5].

### 2.2. Related works

The task of identifying violent or toxic content in the online space has gained much popularity in the NLP domain. Previously, much research focused on the hate speech of social media. Researchers tried to build models to detect hate speech [13] and also tried to analyse the social media comments from various sites like Facebook, Twitter/X, and Reddit.

The rapid rise of Internet users in South Asian countries like Bangladesh has exacerbated issues such as hate speech, particularly in Bengali texts on social media platforms. Despite the prevalence of such content, research on Bengali hate speech detection remains limited. Addressing this gap, our study introduces a novel approach utilizing the BERT architecture combined with a Gated Recurrent Units (GRU) model referred to as GBERT to effectively identify hate speech in Bengali texts [21]. Another work [6] tackles the challenge of generalising hate speech detection algorithms across diverse datasets with unique biases and characteristics. Using an aggregated dataset of 83,230 tweets from 13 different sources, the study evaluated models like TimeLMs, BERTweet, RoBERTa, BERT, and SVM in both binary and multiclass hate speech classification. TimeLMs and BERTweet excelled in single-class classification, while SVM performed best in multiclass scenarios. Despite the strong performance, the study's limitations include a focus on Twitter and English and reliance on base-sized models due to computational constraints.

Despite the widespread use of Bengali on social media, limited research has focused on detecting hate speech in this language. Existing works often lack accuracy and interpretability. Another study proposes [14] an encoder-decoder-based machine learning model to classify Bengali comments from Facebook into seven hate speech categories. Using a dataset of 7,425 Bengali comments, the model employed 1D convolutional layers for feature extraction, with attention mechanism, LSTM, and GRU-based decoders for classification. The attention-based decoder achieved the highest accuracy at 77%, highlighting its potential for improving hate speech detection in Bengali.

In addition, previous works also focus on cyber bullying, abusive comments. The increasing popularity of social media has led to a rise in abusive language and offensive content, particularly challenging to detect in Bengali. Ashraf et al. [7] address this issue by applying Natural Language Processing to classify abusive content using five machine learning classifiers: Decision Tree, Random Forest, Multinomial Naive Bayes, Support Vector Classifier, and Logistic Regression. Preprocessing steps included removing stop words and punctuation, along with tokenization and TF-IDF representation. Logistic Regression achieved the highest accuracy at 85.57%, and the LIME model was employed to enhance model interpretability, providing clarity on the factors influencing predictions. Obaida et al. [26], address the pervasive issue of cyberbullying on social media by proposing a deep-learning model for detection. Utilising three datasets from Twitter, Instagram, and Facebook, the model employs Long Short-Term Memory (LSTM) to identify instances of bullying. The approach demonstrated high effectiveness, achieving accuracies of 96.64%, 94.49%, and 91.26% for the respective platforms. This model represents a significant advancement over previous techniques, successfully addressing key challenges in cyberbullying detection. With the exponential growth of Internet and social media use, cyberbullying has become a significant concern, particularly affecting adolescents. Another study [22] develops a technique for detecting and preventing cyberbullying using Natural Language Processing and machine learning algorithms. Utilising a Kaggle dataset of labelled Twitter data, various classifiers were tested, with Logistic Regression outperforming SVM, Random Forest, Naive Bayes, and XGBoost in terms of accuracy. The ICON-2021 [24] shared task focused on identifying aggression, gender bias, and communal bias across four languages: Meitei, Bengali, Hindi, and English using a dataset of approximately 12,000 comments from social media. Participants had the option to approach the task as separate classifications or a multi-label classification. The top-performing system achieved an overall instance-F1 score of 0.371 and excelled in individual sub-tasks with micro-F1 scores of 0.539 for aggression, 0.767 for gender bias, and 0.834 for communal bias. Despite reasonable performance in individual tasks, achieving accurate three-class classification for aggression proved challenging, with systems correctly classifying only about one-third of instances.

Furthermore, a paper [34] addresses the issue of dataset imbalance in their multi-modal approach for identifying hate speech in Bengali, focusing on 'Sadhu' (সাধু) and 'Cholito' (চলিত) consumption types with neural ensemble models. Their study, involving 30,000 comments, demonstrates that despite the Bengali-Bert model achieving a notable accuracy of 0.706, the imbalance in the dataset remains a significant challenge. They advocate for advanced techniques like over-sampling and synthetic data generation to mitigate the effects of imbalance and enhance model performance.

### 2.3. Research gap

Existing studies indicate that South Asian countries are particularly vulnerable to communal conflicts, yet the focus has been insufficient on how social media-driven content directly contributes to these tensions. While there is considerable research on the role of social media in spreading hate speech and influencing social issues, there remains a critical gap in developing specialised models that detect and analyse comments specifically influencing communal violence. The absence of targeted detection models for communal violence highlights a significant research gap. Addressing this gap could lead to more effective tools for monitoring and mitigating the impact of social media on communal conflicts, ultimately contributing to a more nuanced understanding and management of such violence.

## 3. Dataset

### 3.1. Dataset overview

For our research on detecting communal violent text, we used a dataset that was sourced from the reference paper [33]. This dataset was collected from popular online spaces used by Bengali content creators, focusing on the comments sections of various social media posts. This included diverse identity-based and expression-based strata and incidents of communal violence, as well as Noncommunal violent interactions. After high-level filtering, they retained 13,000 comments from an initial 30,000 samples for annotation. The annotation process was comprehensive and involved a seven-step workflow. An expert team, including social scientists, linguist, and psychologist, conducted the annotations. Each sample underwent two-fold annotation, conflict resolution, subjectivity evaluation by a psychologist, expert board validation, and computational analysis for anomalies.

**Table 1**
Expressions and Operational Definitions of Communal Violence

| Definitions of Classes | |
|---|---|
| Religio Communal | Violence against specific religious groups, often targeting minorities, converts, or non-believers. |
| Ethno Communal | Violence directed against individuals or groups based on their ethnic or communal identity. |
| Nondenominational | Violence targeting individuals based on linguistic identity, geographical and cultural differences, rather than solely ethnic distinctions. |
| Noncommunal | A form of violence that does not fall into the above categories of communal violence. |
| Definitions of Labels | |
| Derogation | Demeaning individuals or groups. |
| Antipathy | A strong negative sentiment towards individuals or groups. |
| Prejudication | Actions or attitudes that support, justify, deny, or falsely accuse misdeeds, mistreatments, and discriminations. |
| Repression | An intent, willingness, or desire to cause harm to others. |

The dataset [33] comprises a total of 12,791 rows and 5 columns. These columns consist of the primary text along with four classes of violence: Religio communal, Ethno communal, Nondenominational communal, and Noncommunal. The violence types are divided into four specific subclasses: 1 (derogation), 2 (antipathy), 3 (prejudication), and 4 (repression). Table 1 provides a clear understanding of the four main classes and their respective sub-classes meanings, as detailed in the original paper [33].

### 3.2. Analysis of dataset

The dataset predominantly comprises non-violent texts and comments, making up 65% of the total data, while the remaining 35% is classified as violent data. Among the violent data, four classes of communal violence are identified. Noncommunal incidents represent 67.8%, Religio communal accounts for 26.4%, Ethno communal for 4.5%, and Nondenominational communal for 1.3% (see Figure 1).

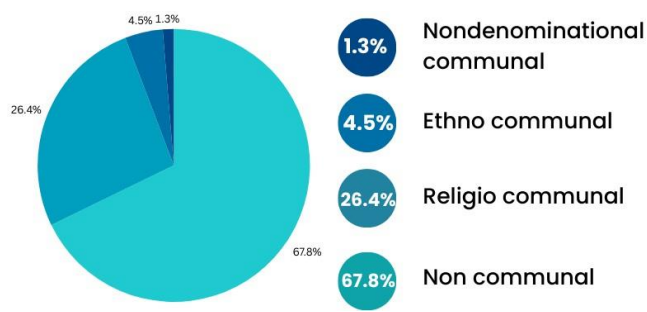

**Figure 1:** Four Class Distribution

The data distribution across sixteen specific classes shows a wide range of frequencies, with categories such as Noncommunal prejudication and derogation having counts of 1,015 and 960 respectively. In contrast, some categories, like nondenominational antipathy and repression, have notably low data, with only 7 and 1 instances, respectively (see Figure 2). This significant imbalance indicates that some classes have so few examples that it is challenging, if not impossible, to work with them effectively.

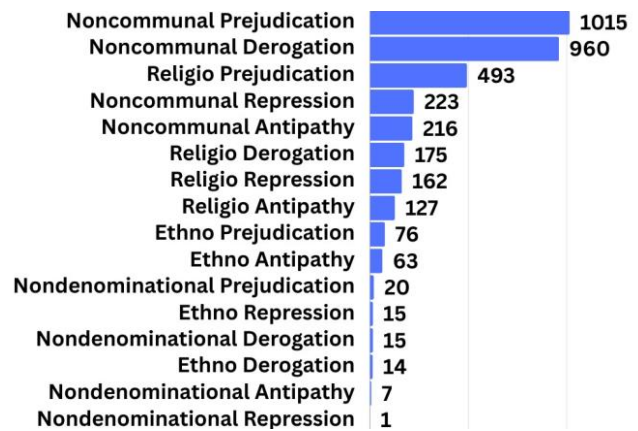

**Figure 2:** Sixteen Class Distribution

### 4. Methodology

We initially focused our analysis on four main classes: Religio communal, Ethno communal, Nondenominational communal, and Noncommunal. However, we encountered significant data scarcity, particularly in the Nondenominational communal and Ethno communal classes, which had very few instances. We also considered expanding our analysis to include sixteen classes, derived from the four main classes with each class having four labels. However, this expansion intensified the data imbalance and scarcity issues, making it infeasible to train effective models across all sixteen classes. As a result, we decided to revert to the original four main classes, simplifying our approach by categorizing each text entry with a 0/1 value for these classes.

The general workflow of this study is presented in Figure 3. Initially, data is prepared through augmentation and preprocessing to enhance and clean the dataset. This refined data is then used to select and optimize the model, ensuring it is well-suited to the task at hand. Finally, the model's performance is evaluated through result analysis and error analysis, which provide insights for further refinement.

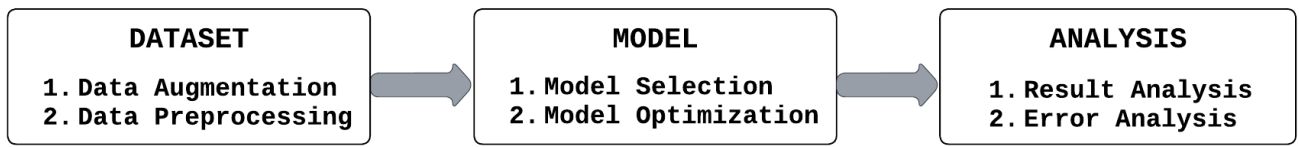

**Figure 3:** General workflow of the study

**Table 2**
SMOTE Examples

| Original Text | Synthetic Text |
|---|---|
| মা দূর্গা যেন এই অসুরদের বিনাশ করেন। (May Goddess Durga destroy these demons.) | মা সকল অসুর যেন দূর্গা এই করেন। (Mother, all demons may Durga do this.) |
| আগে নিজের দেশের হিন্দুদের নিয়ে বলুন। (First, talk about the Hindus in your own country.) | নিজ হিন্দু দেশ থেকে বলুন না। (Do not speak from your own Hindu country) |
| পুলিশ এদেরকে করছ ফায়ার করতে বয় পায়। (The police are afraid to cross-fire them. | এর কর এই কর পা হয় এই। (Doing this, doing this, it becomes.) |

## 4.1. Data augmentation

To address the imbalance within these four classes, we applied five data augmentation methods: SMOTE, Zero-shot, Few-shot, Data paraphrasing, and Manual Data augmentation. This approach allowed us to add more Ethno communal and Nondenominational communal comments which balanced our dataset more effectively.

Synthetic Minority Over-sampling Technique (SMOTE) [12] produced suboptimal results in this context. As demonstrated in the examples (see Table 2), the generated synthetic texts lack coherence and deviate significantly in meaning from the original texts. In essence, when it comes to word embeddings (high-dimensional vectors that contain the information about the meaning and the context of a word) pretrained models like BanglaBERT or mBERT, using SMOTE encounters several limitations [23].

Afterward, we examined Zero-Shot and Few-Shot approaches. Initially, Zero-Shot showed no improvement (see Table 3) despite our intervention [27]. Therefore, we turned to Few-Shot [18], where we used the Z-map [32] method to increase our dataset's volume by mapping features of new samples to the feature space from few-shot samples, Few-Shot also proved ineffective due to a semantic mismatch between source and target data, stemming from the highly contextual nature of the Bengali language (see Table 4).

We then continued the data augmentation process by using Bangla paraphrase model [3] to paraphrase data for the minor classes, Ethno communal and Nondenominational communal, aiming to enhance data variability. However, due to the low initial quantity of data, the paraphrasing process offered minimal benefits (see Table 5). Additionally, we had to manually exclude poorly performed paraphrased data because the model had insufficient text to generate varied and meaningful paraphrases, thus limiting the impact. Consequently, the lack of data remained a significant challenge, preventing the effectiveness of our approach.

**Table 3**
Zero-Shot Examples

| Text | Class Label | Predicted Label |
|---|---|---|
| বাংলাদেশ ইতারারে জাগা দি নিজের ফুন নিজে মাইজ্জে। (Bangladesh fucked themselves by giving these people space to stay.) | Ethno Communal | Unrelated Label |
| সেজন্যই আল্লাহর গজব ঢেলে দিয়েছে চিনের উপরে। (That is why Allah's wrath has been poured out on China.) | Non Denominational | Unrelated Label |

**Table 4**
Few-Shot Examples

| Text | Class Label | Predicted Label |
|---|---|---|
| দেশের এসব লোকেদের ধরে মাইর দেওয়া দরকার। (It is necessary to catch these people and give them a beating.) | Non communal | Ethno communal |
| আল্লাহর গজব পড়ুক তাদের উপরে যারা ইলিশ মাছ বিদেশে দাদা দিদিদের খাওয়ানোর চিন্তা ভাবনা করে। (May Allah's wrath fall upon those who think about sending hilsa fish abroad to their brothers and sisters.) | Non violent | Non denominational |

Due to the ineffectiveness of the above mentioned techniques, we finally opted for annotating new data. We collected comments on YouTube and Twitter that has the potential of causing communal violence, focusing on content that might trigger religious, geopolitical, or other conflicts. Along with our manually curated dataset, we utilized four notable Bengali hate speech datasets obtained from GitHub and Kaggle, totaling 300,000 comments: "Bengali Hate Speech Detection Dataset by UCI" [16],"Bengali Hate Speech Dataset by Nauros from Kaggle" [29],"Multi-Labeled Bengali Toxic Comments"[9],"rezacsedu/Bengali-Hate-Speech-Dataset" [19].

**Table 5**
Paraphrase Examples

| Initial text | Paraphrased text |
|---|---|
| রোহিঙ্গাদের ঠাই না দিলে দেশে আগুন জ্বলে অমানুষের সংখ্যা বেড়ে গেছে। (If the Rohingyas are not given a place, the country will burn and the number of inhuman people has increased.) | রোহিঙ্গাদের যদি ধরা না পড়ে তাহলে দেশে আগুন লেগে যায় মানুষের সংখ্যা বৃদ্ধি পায়। (If the Rohingyas are not caught, then fire breaks out in the country and the number of people increase.) |
| আমরা বাঙ্গালীরা ঝগড়া খুব ভাল পারি। (We Bengalis are very good at arguing.) | আমরা বাঙালিরা খুব ভাল যুদ্ধ করতে পারি। (We Bengalis can do war very well.) |
| নবিহোসেন কে সহ সকল রোহিঙ্গাদেরকে দ্রুত পুশবেক করে দিন। (Including Nabi Hossain, push back all the Rohingyas fast. | নোবিহোসেনসহ সকল রোহিঙ্গাকে দ্রুত তাড়িয়ে দিন। (Quickly expel all Rohingyas, including Nobihossain.) |

Finally, using a fine-tuned BanglaBERT model, we classified these large amounts of comments as communally violent or not. This classification process was followed by manual annotations through a blind voting system. We ensured proper annotations by adhering to the annotation guidelines and definitions of the original paper [33]. To prevent fatigue, we limited annotations to 50 comments per session. Through this process, we added 1,794 new entries, including 1,073 to the Nondenominational communal class, 508 to the Ethno communal class, and 300 to the Religio communal class. This augmentation helped balance the dataset, significantly increasing the proportion of Ethno communal data and Nondenominational communal data. Specifically, the dataset saw a considerable rise in Nondenominational communal data, which expanded from just 1.2% to 7.2%, while Ethno communal data doubled from 4.69% to 8.6% (see Figure 4).

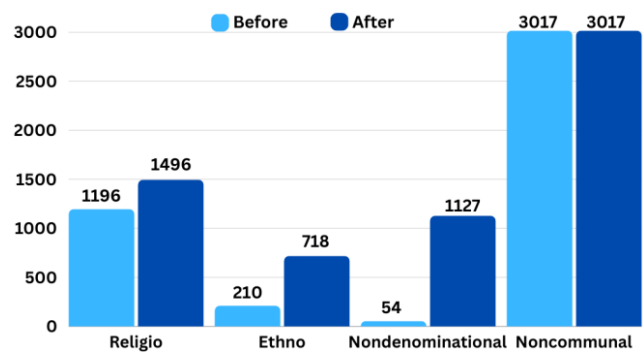

**Figure 4:** Impact of Manual Data Augmentation

### 4.2. Data preprocessing

We carried out multiple preprocessing steps to improve the quality of the Bengali text comments. First, we removed stop words [17] to improve clarity and relevance. Next, we fixed broken characters to ensure readability and preserve linguistic integrity. We removed linguistic noise by eliminating punctuation, unnecessary spaces, and garbage characters. Emojis were replaced with their text meanings using Kaggle's "Emoji_Dict.p" [2] to preserve expressive nuances (e.g., "🥺" became ":pleading_face_emoji:"). This maintained the original sentiments for NLP analysis. Initially, we divided our dataset into 80% train and 20% test data. Then, we further split the training data into 85% for training and 15% for validation. Finally, we used the "csebuetnlp/banglabert" [10] tokenizer with a maximum length of 512, chosen for its optimization for Bengali language features.

### 4.3. Model

In this study, we utilised three pre-trained bert models: BanglaBERT, BanglaBERT Large, and Multilingual BERT (mBERT). Our initial approach followed the baseline model as outlined in the original paper [33], which used BanglaBERT. We employed BanglaBERT with the same hyperparameters specified in the baseline to maintain consistency. Subsequently, we fine-tuned the model to enhance its performance further. BanglaBERT [10] is a specialised pre-trained language model tailored explicitly for the Bengali language. It addresses the challenges associated with natural language processing (NLP) in low-resource languages by leveraging large-scale Bengali corpora for pre-training.

For tokenization, we utilised the BERT auto tokenizer with a maximum sequence length of 512 tokens. To prevent overfitting, we implemented early stopping after three epochs. This approach was combined with a best-saved model strategy, where model checkpoints were saved periodically, and the best-performing model on the validation set was selected for final deployment. These strategies ensured optimal performance and generalisation capabilities.

To further refine the performance of our classification model, we transitioned to using BanglaBERT Large [10], a more extensive variant of BanglaBERT with increased parameters and deeper layers. This model was configured with the same hyperparameters as BanglaBERT to maintain a controlled experimental environment. The expanded architecture of BanglaBERT Large is designed to capture more intricate linguistic nuances, which can be especially beneficial in capturing complex syntactic and semantic relationships in the Bengali language.

Additionally, we employed Multilingual BERT (mBERT) to evaluate its effectiveness compared to our Bengali-specific models and to investigate any anomalies in our results. mBERT [15], an extension of Google's BERT, is capable of understanding and generating text across multiple languages simultaneously. Its multilingual training corpus allows it to generate universal representations that are transferable across languages, including Bengali. This cross-lingual transfer learning is particularly advantageous for low-resource languages like Bengali, where annotated datasets are often limited. By incorporating mBERT, we aimed to benchmark the performance of our Bengali-specific models against a widely recognized multilingual model, providing insights into the generalizability and effectiveness of language-specific versus multilingual approaches in the context of Bengali NLP tasks.

## 5. Result analysis

In this study, we analyzed the performance of fine-tuned models in two setups: one for four-class classification and another using ensemble methods. For the four-class classification task, we conducted a more exhaustive study, training several models based on the hyperparameters tuned on an optimized set. This optimized set was determined through Bayesian optimization, which efficiently handles parameters like learning rate and batch size. Additionally, we utilized class weights and data augmentation techniques. The best fine-tuned models from the four-class setting were then used to create an ensemble model, combining their optimized strengths to enhance overall performance.

### 5.1. Four class metrics

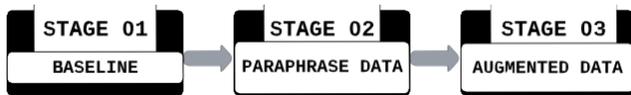

**Figure 5:** Stages of four class metrics

For the four-class classification task, we proceed through three stages sequentially (see Figure 5). Initially, we observed that the dataset had a significantly low proportion of Ethno communal and Nondenominational communal data. This hampered the baseline model to accurately distinguish between these specific classes (see Table 6). So, we added reciprocal class weight so that these two minority classes get more priority and got a considerable boost in F1 scores at Religio communal and Nondenominational communal, but a decrease in the other two classes.

To further explore the impact of adding class weights, we incorporated paraphrased data generated using the Bangla-paraphrase model [3]. However, the results were not satisfactory due to the poor quality of the paraphrased data. While F1 scores improved for all classes except Religio communal, a slight decrease in this class from 0.55 to 0.53 could be attributed to over-generalization from the paraphrased data. On the other hand, the notable improvements in Ethno communal from 0.32 to 0.48 and Nondenominational communal from 0.16 to 0.29 highlight the advantages of data augmentation.

Following the incorporation of manual data augmentation alongside paraphrased data, we observed an overall improvement in F1 scores. However, this enhancement also exposed a conflict between the Religio communal and Noncommunal data. To address the data imbalance, we specifically targeted paraphrasing for the Ethno communal and Nondenominational communal classes. Despite these efforts, the quality of the paraphrased data was insufficient, leading to suboptimal results. Consequently, to mitigate these issues and improve model performance, we decided

**Table 6**
Performance Metrics of Fine-tuned Models

| Class | Prec. | Recall | F1 | Macro F1 |
|---|---|---|---|---|
| **BanglaBert Baseline [No Class Weight]** | | | | |
| Religio-communal | 0.48 | 0.50 | 0.49 | |
| Ethno-communal | 0.50 | 0.35 | 0.41 | 0.35 |
| Nondenominational | 0.00 | 0.00 | 0.00 | |
| Noncommunal | 0.56 | 0.48 | 0.52 | |
| **BanglaBert Baseline** | | | | |
| Religio-communal | 0.47 | 0.65 | 0.55 | |
| Ethno-communal | 0.21 | 0.72 | 0.32 | 0.37 |
| Nondenominational | 0.11 | 0.32 | 0.16 | |
| Noncommunal | 0.60 | 0.34 | 0.44 | |
| **BanglaBert [Paraphrased Data]** | | | | |
| Religio-communal | 0.45 | 0.66 | 0.53 | |
| Ethno-communal | 0.39 | 0.60 | 0.48 | 0.43 |
| Nondenominational | 0.23 | 0.37 | 0.29 | |
| Noncommunal | 0.61 | 0.35 | 0.44 | |
| **BanglaBert [Paraphrased & Augmented Data]** | | | | |
| Religio-communal | 0.54 | 0.41 | 0.46 | |
| Ethno-communal | 0.55 | 0.77 | 0.64 | 0.49 |
| Nondenominational | 0.38 | 0.85 | 0.53 | |
| Noncommunal | 0.68 | 0.22 | 0.34 | |
| **BanglaBert [Augmented Data & No Class Weights]** | | | | |
| Religio-communal | 0.55 | 0.39 | 0.46 | |
| Ethno-communal | 0.46 | 0.72 | 0.56 | 0.53 |
| Nondenominational | 0.51 | 0.76 | 0.61 | |
| Noncommunal | 0.62 | 0.38 | 0.47 | |
| **BanglaBert [Augmented Data]** | | | | |
| Religio-communal | 0.51 | 0.43 | 0.47 | |
| Ethno-communal | 0.61 | 0.72 | 0.66 | **0.60** |
| Nondenominational | 0.86 | 0.58 | 0.69 | |
| Noncommunal | 0.54 | 0.60 | 0.57 | |
| **BanglaBert Large [Augmented Data]** | | | | |
| Religio-communal | 0.42 | 0.60 | 0.49 | |
| Ethno-communal | 0.48 | 0.77 | 0.59 | 0.51 |
| Nondenominational | 0.46 | 0.89 | 0.61 | |
| Noncommunal | 0.73 | 0.23 | 0.35 | |
| **mBert [Augmented Data]** | | | | |
| Religio-communal | 0.30 | 0.72 | 0.42 | |
| Ethno-communal | 0.51 | 0.76 | 0.61 | 0.52 |
| Nondenominational | 0.70 | 0.64 | 0.67 | |
| Noncommunal | 0.48 | 0.33 | 0.39 | |

**Hyperparameters**: Epochs - **30** with **Early stopping** including **patience size of 2**, Batch Size - **32**, Learning rate - **2e-5**.

to proceed with data augmentation exclusively for subsequent model iterations. In the manual augmented data, we evaluated two approaches: one with class weights and one without. Given the imbalance in our dataset, the approach utilizing class weights demonstrated better results.

With manual data augmentation, the model's performance improved significantly, with the F1 score for Nondenominational communal rising from 0.29 to 0.69, for Ethno-communal increasing from 0.48 to 0.66, compared to the results with paraphrased data. To mitigate the class imbalance after augmentation, we applied inverse frequency class weights to enhance the impact of the underrepresented classes: Religio communal, Ethno communal, and Nondenominational communal, as they were less prevalent compared to the Noncommunal class. This adjustment led to significant improvements in F1 scores across all classes, indicating enhanced model sensitivity and precision for minority classes. The model achieved the highest Macro F1 score of 0.60. Despite these improvements, the conflict between the Religio comunal and Noncommunal classes remained. To address this ongoing issue, we decided to switch to the BanglaBERT large model for future work, expecting to take advantage of its advanced capabilities to better resolve these conflicts.

Using the BanglaBERT large model, we discovered that overall F1 scores were lower across all classes than in our previous model, with a maximum Macro F1 score of 0.51. Furthermore, the conflict between the Religio comunal and Noncommunal classes remained indicating that this model did not solve the problem. To further address this challenge, we switched to the mBERT model, a multilingual model, to see if it could provide better performance. However, this switch produced F1 scores that were even lower than our best-performing model, BanglaBERT, with a maximum Macro F1 score of only 0.52. These findings indicate that neither BanglaBERT large nor mBERT effectively addressed classification conflicts or improved overall performance.

## 5.2. Ensemble model

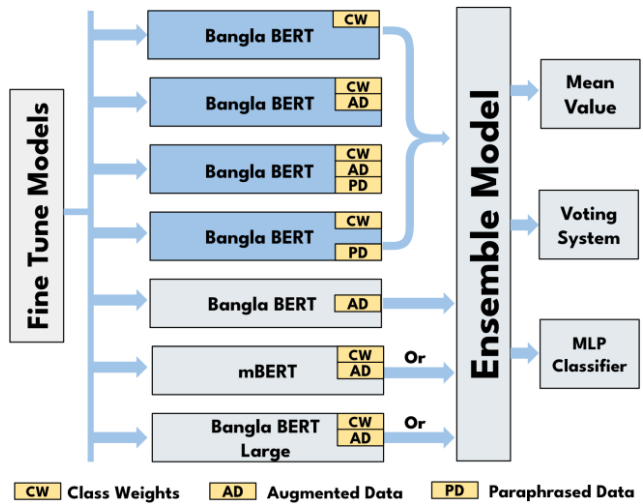

**Figure 6:** Ensemble Model Approach

For communal violence classification tasks, we applied ensemble approaches to increase classification performance using multiple fine-tuned BERT models. Ensemble approaches improve the accuracy and reduce the inaccuracy of individual models by combining predictions from different models. Specifically, we explored three types of ensemble techniques: The three methods are: the mean value approach, the voting system, and the MLP classifier [35].

**Table 7**
Performance Metrics of Ensemble Models

| Techniques | Class | F1 | Macro F1 |
|---|---|---|---|
| **Ensemble Model-1** | | | |
| Mean Value | Religio-communal | 0.562 | 0.61 |
| | Ethno-communal | 0.656 | |
| | Nondenominational | 0.782 | |
| | Noncommunal | 0.461 | |
| Voting System | Religio-communal | 0.552 | 0.61 |
| | Ethno-communal | 0.662 | |
| | Nondenominational | 0.758 | |
| | Noncommunal | 0.479 | |
| MLP Classifier | Religio-communal | 0.463 | 0.60 |
| | Ethno-communal | 0.681 | |
| | Nondenominational | 0.744 | |
| | Noncommunal | 0.512 | |
| **Ensemble Model-2** | | | |
| Mean Value | Religio-communal | 0.573 | 0.62 |
| | Ethno-communal | 0.671 | |
| | Nondenominational | 0.781 | |
| | Noncommunal | 0.474 | |
| Voting System | Religio-communal | 0.559 | 0.62 |
| | Ethno-communal | 0.673 | |
| | Nondenominational | 0.752 | |
| | Noncommunal | 0.481 | |
| MLP Classifier | Religio-communal | 0.453 | 0.60 |
| | Ethno-communal | 0.712 | |
| | Nondenominational | 0.744 | |
| | Noncommunal | 0.513 | |
| **Ensemble Model-3** | | | |
| Mean Value | Religio-communal | 0.561 | 0.63 |
| | Ethno-communal | 0.662 | |
| | Nondenominational | 0.773 | |
| | Noncommunal | 0.512 | |
| Voting System | Religio-communal | 0.571 | **0.63** |
| | Ethno-communal | 0.671 | |
| | Nondenominational | 0.763 | |
| | Noncommunal | 0.511 | |
| MLP Classifier | Religio-communal | 0.471 | 0.61 |
| | Ethno-communal | 0.691 | |
| | Nondenominational | 0.742 | |
| | Noncommunal | 0.522 | |

In our research, we used an ensemble model that consisted of five fine-tuned models. We used four distinct variations of BanglaBERT: (1) BanglaBERT with added class weights, (2) BanglaBERT with paraphrased data and class weights, (3) BanglaBERT with augmented and paraphrased data with class weights, (4) BanglaBERT with augmented data without class weights. For the ensemble's fifth model, we created three configurations (see Figure 6): **Ensemble Model-1**: The four BanglaBERT variations plus BanglaBERT Large (Augmented Data with Class Weights). **Ensemble Model-2**: The four Bangla BERT variations plus mBERT (Augmented Data with Class Weights). **Ensemble Model-3**: The four BanglaBERT variations plus another instance of BanglaBERT (Augmented Data with Class Weights).

These configurations resulted in three types of ensemble combinations, which we denoted as **Ensemble Model-1**, **Ensemble Model-2**, and **Ensemble Model-3** in Table 7.

In our individual fine-tuned models, we encountered a difficult conflict between Religio communal and Noncommunal data, an anomaly that frequently impacted accuracy in prediction (see Table 6). However, through the process of ensemble learning, we have been able to considerably decrease this conflict, which contributed to an overall increase in the Macro F1 score compared to our best-performing individual model (BanglaBERT with Augmented Data). Among the ensemble models, Ensemble Model 3's voting system yielded the best macro F1 score.

The F1 score for the Religio communal class improved significantly from 0.47 in the best-performing individual model to 0.57 in the best performing Ensemble model. For the Nondenominational class, it increased from 0.69 to 0.77. These gains were pivotal in driving the overall increase in the Macro F1 score. Although there was a slight decline in the F1 score for the Noncommunal class, which dropped from 0.57 to 0.51, the ensemble approach proved highly effective in enhancing the accuracy of class-specific predictions.

Among the ensemble techniques used, both the mean value and voting system yielded the same F1 scores. However, the voting system has a slight edge over the mean value (see Table 7), primarily because it minimizes the effect of the outliers or extreme predictions, ensuring that the final decision reflects the majority view [1]. This is especially valuable in classification problems, as it generates straightforward and easily interpretable results if individual models are diverse or flawed. Additionally, the voting system proved more adaptive to class imbalances, enhancing predictions in minority classes like Religio communal and Ethno communal, leveraging the collective learning strength of **Ensemble Model-3**. This ensemble method has delivered the most accurate confusion matrix results (see Figure 7) so far.

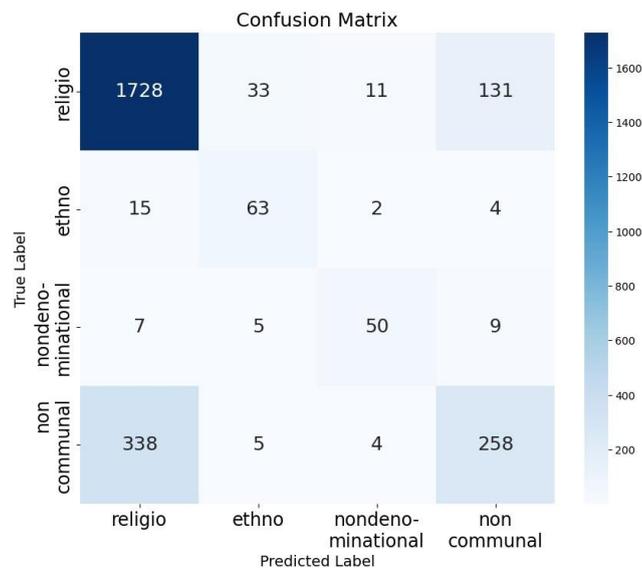

**Figure 7:** Conf. Matrix of Ensemble Model-3 [Voting System]

## 6. Error analysis
### 6.1. Data annotation anomalies and challenges

A significant challenge in the dataset was incorrect data annotation, impacting our model's performance. For analysing this, we collected the misclassified texts from our best-performing model, randomly selected 300 texts, and manually reviewed them. We found 54 problematic annotated texts, which accounts for 18% of the total reviewed texts. Some of these texts were clearly mislabeled (see Table 8), while some were difficult to classify due to the lack of context (see Table 9).

### 6.2. Pre-trained model limitations
#### 6.2.1. BanglaBERT

We performed an analysis by examining the frequent words in both the Noncommunal and Religio communal classes (see Table 10). In the Noncommunal class, we noticed that many frequent words were quite generic, often consisting of stop words or words that lacked substantial semantic meaning on their own such as ভাই (Brother), ভালো (Good), and লজ্জা (Shame). However, we identified two meaningful words, মানুষ (Human) and বাংলাদেশ (Bangladesh), that stood out from the Noncommunal class. On the other hand, in the Religio communal class, the frequent words were highly relevant and specific to religious contexts. These included words like কাফের (Infidel), নাস্তিক (Atheist), ধর্ম (Religion), মুসলিম (Muslim), ইসলাম (Islam), and আল্লাহ (Allah), which are critical for identifying religious themes and sentiments. To understand why the BanglaBERT model was struggling with misclassification, we calculated the cosine similarity between the word vectors of the selected Noncommunal words and the frequent Religio communal words. In natural language processing, word embeddings are used to represent words as high-dimensional vectors, capturing their meanings based on the context in which they appear. Cosine similarity measures the cosine of the angle between two vectors, indicating how similar they are in terms of direction. In our analysis, we found that the cosine similarity between the Noncommunal words and the Religio communal words was unexpectedly high. This means that, in the vector space, these Noncommunal words are positioned close to the Religio communal words, suggesting that the model perceives them as being contextually similar. Because of this high similarity, the model has difficulty distinguishing between the two classes, leading to frequent misclassifications. Essentially, the model is tricked by the closeness in the vector space, mistaking Noncommunal text for Religio communal content due to the overlapping semantic features captured by the word embeddings.

#### 6.2.2. BanglaBERT Large

In the BanglaBERT Large model, the cosine similarity between words in the Religio communal and Noncommunal classes was even higher than in the standard BanglaBERT model. This increased similarity exacerbated the model's difficulty in distinguishing between the two classes, as the words appeared even closer in the vector space. As a result, the model struggled more with classification, leading to a significant drop in performance. Specifically, the F1 score for Noncommunal classes decreased by 0.16 compared to the best-performing standard BanglaBERT model (see Table 10).

**Table 8**
Overview of Misannotations vs. Model Predictions Across Four Classes

| Text | Annotation | Model Pred. | Pred. Score |
|---|---|---|---|
| প্রবাদের একটি নতুন সংস্করণ হলো 'খাল কেটে রোহিঙ্গা আনা'। (A new version of the proverb is `Digging a canal to bring Rohingyas') | Non Violent | Ethno Communal | 0.9401 |
| হিজাব আর বোরখা ছাড়া নারী মানেই মোল্লাদের চোখে উলংগ নারী..... (A woman without burkha and hijab is a naked woman in the eyes of mullahs.) | Non Violent | Religio Communal | 0.7581 |
| তোকে নাস্তিক-নাস্তিক সন্দেহ হয়... তোকে বেজন্মা-বেজন্মা মনে হয়... তোর জন্ম ঠিক জায়গায় হয়নি। (I suspect you are an atheist... it seems you are born out of wedlock... you were not born in the right place.) | Non Violent | Religio Communal | 0.7004 |
| ধর্ম-ব্যবসায়ী, ধর্মান্ধ, জংগী ও উগ্রবাদীদের ইসরায়েলের বিরুদ্ধে জিহাদ ঘোষণা করা উচিত। (Religious profiteers, Extremists, fanatics, and radicals should declare jihad | Noncommunal | Religio Communal | 0.6965 |
| মন্ত্রী মহাশয়, আপনি দায়িত্ব নিয়ে ইসরায়েলকে দমন করুন। (Take on a responsibility and suppress Israel, honorable minister.) | Non Violent | Noncommunal | 0.6301 |

**Table 9**
Overview of context lacking text

| Text | Annotation | Model Pred. |
|---|---|---|
| এক যে ছিলো শিয়ালে, মোরগ আঁকে দেয়ালে, আপন মনে চাটতে থাকে খেয়ালে। (Once there was a fox, who drew a rooster on the wall, and kept licking it in its own thought.) | Noncommunal | Non Violent |
| সমস্যা ১০০% সমাধান না হওয়া পর্যন্ত কাউকেই ধন্যবাদ দেওয়া যাবে না! সবই চীন ভারতের বিজনেস! (No one can be thanked until the problem is 100% solved! Everything is China-India business!) | Noncommunal | Non Violent |
| করোনাভাইরাস আশার কারন নির্যাতন। (The reason for the coronavirus is oppression.) | Noncommunal | Non Violent |
| ভাগ্য ভালো বাংলাদেশে জন্ম গ্রহণ করছি না হলে কতো বিনোদন যে মিস করতাম। (Fortunate that I was born in Bangladesh; otherwise, how much entertainment I would have missed.) | Noncommunal | Non Violent |
| আপদ বিপদ ডেকে নিয়ে আসছি আমরা দেশে। (We are inviting calamities and dangers into the country.) | Ethno Communal | Non Violent |

**Table 10**
Cosine Similarity: Religio Vs Noncommunal Common Words

| Non-communal Word | Religio-Communal Word | Bangla BERT | Bangla BERT Large | mBERT |
|---|---|---|---|---|
| মানুষ (Human) | কাফের (Infidel) | 0.9539 | 0.9419 | 0.4839 |
| | নাস্তিক (Atheist) | 0.9706 | 0.9703 | 0.6049 |
| | ধর্ম (Religion) | 0.9701 | 0.9873 | 0.6591 |
| | মুসলিম (Muslim) | 0.8825 | 0.9789 | 0.5854 |
| | ইসলাম (Islam) | 0.9631 | 0.9358 | 0.6146 |
| | আল্লাহ (Allah) | 0.9413 | 0.9591 | 0.5444 |
| | হিন্দু (Hindu) | 0.9178 | 0.9678 | 0.6154 |
| বাংলাদেশ (Bangladesh) | কাফের (Infidel) | 0.8966 | 0.8978 | 0.4907 |
| | নাস্তিক (Atheist) | 0.9282 | 0.9233 | 0.4195 |
| | ধর্ম (Religion) | 0.9429 | 0.9494 | 0.4209 |
| | মুসলিম (Muslim) | 0.9218 | 0.9674 | 0.5267 |
| | ইসলাম (Islam) | 0.9340 | 0.9848 | 0.4656 |
| | আল্লাহ (Allah) | 0.9548 | 0.9843 | 0.4234 |
| | হিন্দু (Hindu) | 0.9365 | 0.9623 | 0.5614 |

### 6.2.3. mBERT

In the case of mBERT, the cosine similarity issue was not as severe as in the other models. For instance, the cosine similarity score between আল্লাহ (Allah) and বাংলাদেশ (Bangladesh) was more balanced at 0.4234. However, despite this, mBERT struggled with accurately classifying Religio communal and Noncommunal texts, achieving F1 scores of just 0.42 and 0.39, respectively. The primary challenge for mBERT was its difficulty in capturing the subtle nuances and contextual details of the Bengali language. As a multilingual model, mBERT is trained in a broad range of languages, which may cause it to miss specific cultural and linguistic details crucial for understanding Bengali text. While mBERT performed well in other categories, its broad training scope limited its ability to grasp the fine-grained distinctions necessary for accurately differentiating between Religio communal and Noncommunal content in Bengali (see Table 10).

### 6.3. Our models' limitation

As mentioned before, our models struggled to distinguish between Religio communal and Noncommunal texts.

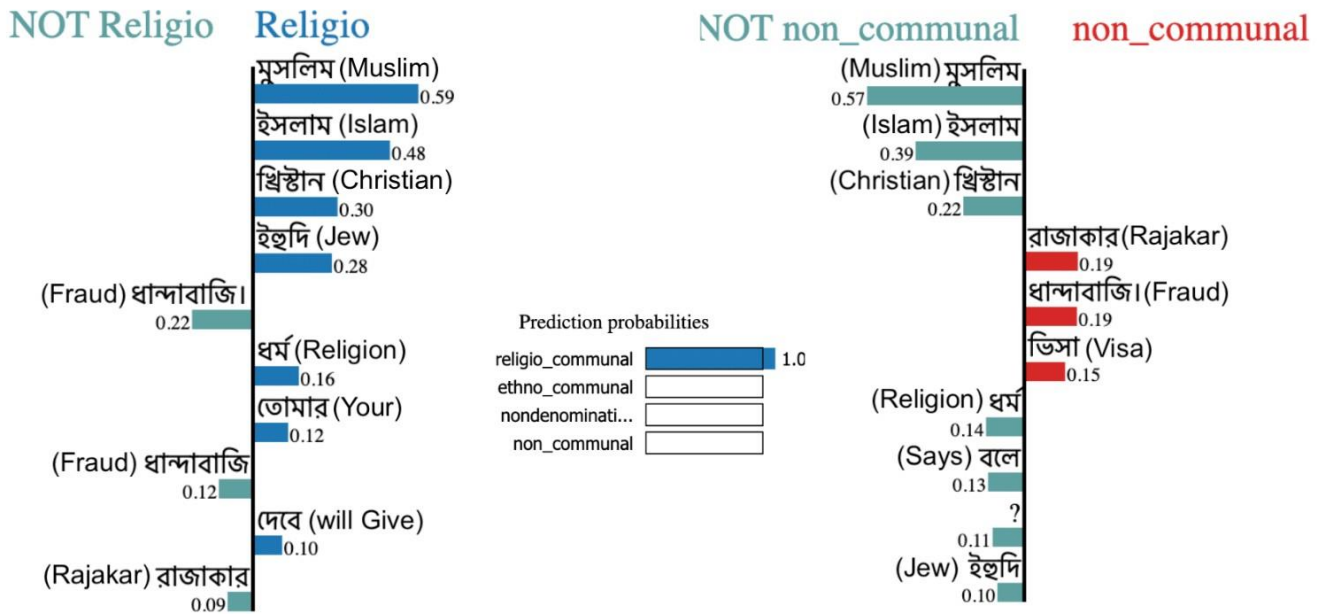

**Figure 8:** LIME Analysis - Example 1

The potential reason for that is the presence of significant words with religious context in many Noncommunal texts.

Through LIME analysis, we identified the top 10 words that have the highest weights in influencing the classification of texts as Religio communal. These words were found to heavily impact the model's decisions. These words are হিন্দু (Hindu), কাফের (Infidel), মুসলিম (Muslim), হিন্দুদের (of Hindus), ইসলাম (Islam), কোরআন (Quran), ইসলামের (of Islam), মুসলমানদের (Muslims), ধ্বংস (destruction), and আল্লাহর (of Allah). We analyzed the model's misclassifications by focusing on Noncommunal texts that were incorrectly labeled as Religio communal by the model. Our findings revealed that 65% of these misclassified Religo communal texts contained the aforementioned words. This suggests that the presence of these strongly religious words in Noncommunal texts may be a significant factor contributing to their misclassification as Religio communal. Using LIME analysis, we also examined individual misclassified texts to understand the role of specific words in these misclassifications. In this example (see Figure 8), terms like মুসলিম (Muslim), ইসলাম (Islam), and খ্রিস্টান (Christian) were positively weighted towards the Religio communal class, strongly influencing the model to predict Religio communal. Conversely, these same words had negative weights in the Noncommunal class, decreasing the likelihood of a Noncommunal classification. As a result, the model predicted the text as Religio communal with high confidence.

We extended our analysis to understand the model's confusion in identifying Religio communal texts misclassified as Noncommunal. We observed that some words such as গজব (God's Curse), ধ্বংস (Destruction), and মারা (Kill) have been assigned positive weights towards both Noncommunal and Religio communal classes. This crossover makes it difficult for the model to distinguish and assign suitable weights to these words. In this example (see Figure 9), the word গজব (God's Curse) was heavily weighted in both the Religio communal and Noncommunal class, causing the misclassification of the text. These overlapping influences of religious terms and ambiguous weights are key reasons for the model's misclassification between Religio communal and Noncommunal texts.

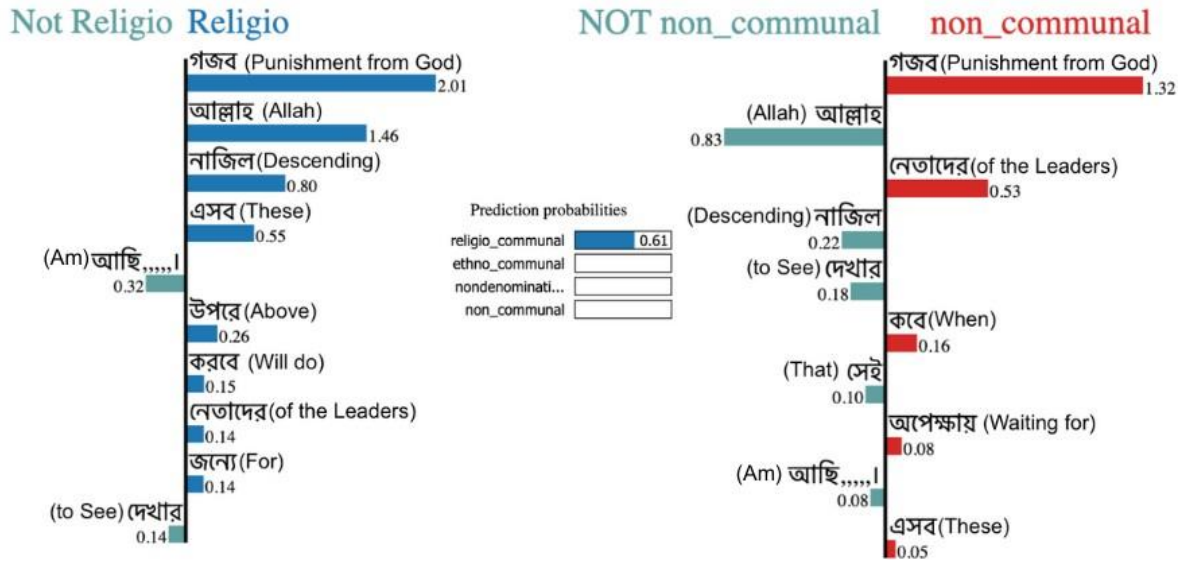

**Figure 9:** LIME Analysis - Example 2

## 7. Future work

### 7.1. Reducing class imbalances

Due to a lack of data for analysis, we only looked at four classes of communal violent text in this study. Future research should focus on raising the dataset's size to include enough samples for all sixteen classes, which we failed to do due to a significant class imbalance, with noncommunal content accounting for one-third of the whole dataset. Annotation issues were also found, with some annotations being incorrect and others lacking proper context. With more extensive and balanced data, as well as correct annotations, the model can be trained more comprehensively, hence improving its performance and reliability.

### 7.2. Model improvement based on error Analysis

Given the findings of our error analysis, future research will focus on developing techniques to improve the model in areas where it makes mistakes. For example, in our present fine-tuned model, Religio communal and Noncommunal violence are frequently confused due to BanglaBERT models' high cosine similarity among relevant words. Future improvements will include increasing the classification accuracy between these classes, which may require more sophisticated techniques such as contextual embedding and domain specific fine tuning.

## 8. Conclusion

Our research focused on the relatively under-researched area of hate speech in the Bengali language, specifically aiming to understand and analyze communal violent text on social media using large language models. We expanded our dataset by adding 1,794 new data points in the classes of Religio communal, Ethno communal, and Nondenominational communal, which supported our hypothesis that a more diverse, multi-class dataset would help the model better recognize and classify various forms of violent text. We were able to improve the F1 score from 0.36 in the initial baseline model to 0.63 by augmenting data and developing an enhanced Ensemble Model. Our fine-tuning of BanglaBERT, as hypothesized, significantly improved its performance in classifying communal violence. We also found that while our model generally performed well in distinguishing Noncommunal comments, there were instances of misclassification, particularly when comments overlapped between different classes like Religio communal versus Noncommunal. This was expected, as we had anticipated that real-world comments might be ambiguous and require morecontext to be accurately understood. Overall, our work has made a meaningful contribution to the field of Natural Language Processing by offering a valuable tool that can help identify and address communal violence in Bengali text on social media. We hope that this research will encourage further studies on this important topic and help advance the understanding of violent text detection in the Bengali language.

How Effectively Can BERT Models Interpret Context and Detect Bengali Communal Violent Text?## 9. Ethical Consideration

This study includes sensitive language and names for research purposes only, aiming to improve violent comment detection without causing harm or offense. We are committed to respecting privacy and not demeaning any individuals or groups.

## References

[1] Aeeneh, S., Zlatanov, N., Yu, J., 2024. New bounds on the accuracy of majority voting for multiclass classification. IEEE Transactions on Neural Networks and Learning Systems , 1–15doi:10.1109/TNNLS.2024.3387544.

[2] Agrawal, D., 2020. Emoji dictionary. URL: https://www.kaggle.com/divyansh22/emoji-dictionary-1.

[3] Akil, A., Sultana, N., Bhattacharjee, A., Shahriyar, R., 2022. BanglaParaphrase: A high-quality Bangla paraphrase dataset, Association for Computational Linguistics, Online only. pp. 261–272. URL: https://aclanthology.org/2022.aacl-short.33.

[4] Al Jazeera, 2021a. Hundreds protest in bangladesh over deadly religious violence. URL: https://aje.io/7ndrbq. retrieved January 23, 2024.

[5] Al Jazeera, 2021b. Two hindu men killed, temples vandalised in bangladesh violence. URL: https://aje.io/uy9gmz. retrieved February 18, 2024.

[6] Antypas, D., Camacho-Collados, J., 2023. Robust hate speech detection in social media: A cross-dataset empirical evaluation. URL: https://arxiv.org/abs/2307.01680, arXiv:2307.01680.

[7] Ashraf, K., Hosen, M.H., Asgar, S., Islam, M.T., Nawar, S., 2024. Analyzing abusive bangla comments on social media: Nlp & explainable ai, pp. 1–6. doi:10.1109/iCACCESS61735.2024.10499547.

[8] Bashar, I., 2011. Bangladeshs forgotten crisis: Land, ethnicity, and violence in chittagong hill tracts. Counter Terrorist Trends and Analyses 3, 1–5. URL: http://www.jstor.org/stable/26350972.

[9] Belal, T.A., Shahariar, G.M., Kabir, M.H., 2023. Interpretable multi labeled bengali toxic comments classification using deep learning, pp. 1–6. doi:10.1109/ECCE57851.2023.10101588.

[10] Bhattacharjee, A., Hasan, T., Ahmad, W., Mubasshir, K.S., Islam, M.S., Iqbal, A., Rahman, M.S., Shahriyar, R., 2022. BanglaBERT: Language model pretraining and benchmarks for low-resource language understanding evaluation in Bangla, Association for Computational Linguistics, Seattle, United States. pp. 1318–1327. URL: https://aclanthology.org/2022.findings-naacl.98.

[11] Brosché, J., Elfversson, E., 2012. Communal conflict, civil war, and the state: Complexities, connections, and the case of sudan. African Journal on Conflict Resolution 12, 33–60. URL: https://www.ajol.info/index.php/ajcr/article/view/78700.

[12] Brownlee, J., 2021. Smote for imbalanced classification with python. URL: https://machinelearningmastery.com/smote-oversampling-for-imbalanced-classification/.

[13] Cano Basave, A.E., He, Y., Liu, K., Zhao, J., 2013. A weakly supervised Bayesian model for violence detection in social media, Asian Federation of Natural Language Processing, Nagoya, Japan. pp. 109–117. URL: https://aclanthology.org/I13-1013.

[14] Das, A.K., Asif, A.A., Paul, A., Hossain, M.N., 2021. Bangla hate speech detection on social media using attention-based recurrent neural network. Journal of Intelligent Systems 30, 578–591. doi:10.1515/jisys-2020-0060.

[15] Devlin, J., Chang, M., Lee, K., Toutanova, K., 2018. BERT: pre-training of deep bidirectional transformers for language understanding. CoRR abs/1810.04805. URL: http://arxiv.org/abs/1810.04805.

[16] Dey, K., Sumon, Cochez, M., Karim, M.R., 2022. Bengali Hate Speech Detection Dataset. UCI Machine Learning Repository. URL: https://doi.org/10.24432/C5PD07.

[17] Haque, R.U., Mridha, M.F., Hamid, M.A., Abdullah-Al-Wadud, M., Islam, M.S., 2020. Bengali stop word and phrase detection mechanism. Arabian Journal for Science and Engineering 45, 33553368. doi:10.1007/s13369-020-04388-8.

[18] IBM, 2024. Few-shot learning. URL: https://www.ibm.com/topics/few-shot-learning. retrieved January 29, 2024.

[19] Karim, M.R., Chakravarthi, B.R., McCrae, J.P., Cochez, M., 2020. Classification benchmarks for under-resourced bengali language based on multichannel convolutional-lstm network. URL: https://arxiv.org/abs/2004.07807, arXiv:2004.07807.

[20] Kemp, S., 2023. Digital 2023: Bangladesh. URL: https://datareportal.com/reports/digital-2023-bangladesh.

[21] Keya, A.J., Kabir, M.M., Shammey, N.J., Mridha, M.F., Islam, M.R., Watanobe, Y., 2023. G-bert: An efficient method for identifying hate speech in bengali texts on social media. IEEE Access 11, 79697–79709. doi:10.1109/ACCESS.2023.3299021.

[22] Khang Hsien, Y., Arabee Abdul Salam, Z., Kasinathan, V., 2022. Cyber bullying detection using natural language processing (nlp) and text analytics, in: 2022 IEEE International Conference on Distributed Computing and Electrical Circuits and Electronics (ICDCECE), pp. 1–4. doi:10.1109/ICDCECE53908.2022.9792931.

[23] Kim, M., 2023. Smote: Practical consideration and limitations. URL: https://medium.com/@minjukim023/smote-practical-consideration-limitations-f0d926b661a8.

[24] Kumar, R., Ratan, S., Singh, S., Nandi, E., Devi, L.N., Bhagat, A., Dawer, Y., Lahiri, B., Bansal, A., 2021. ComMA@ICON: Multilingual gender biased and communal language identification task at ICON-2021, NLP Association of India (NLPAI), NIT Silchar. pp. 1–12. URL: https://aclanthology.org/2021.icon-multigen.1.

[25] Noorana, M., 2018. Pre electoral violence in 2001 parliamentary elections in bangladesh: Empirical evidence from bagerhat and narayanganj districts, social science review: The dhaka university studies 35, 231–246. URL: https://www.researchgate.net/publication/353636999.

[26] Obaida, M.H., Elkaffas, S.M., Guirguis, S.K., 2024. Deep learning algorithms for cyber-bulling detection in social media platforms. IEEE Access 12, 76901–76908. doi:10.1109/ACCESS.2024.3406595.

[27] Polat, G., 2023. Zero-shot learning (zsl) explained. URL: https://encord.com/blog/zero-shot-learning-explained/.

[28] ReliefWeb, 2017. Myanmars religious and ethnic conflicts: No end in sight. URL: https://reliefweb.int/report/myanmar/myanmars-religious-and-ethnic-conflicts-no-end-sight. retrieved January 17, 2024.

[29] Romim, N., Ahmed, M., Talukder, H., Islam, M.S., 2020. Hate speech detection in the bengali language: A dataset and its baseline evaluation. arXiv preprint arXiv:2012.09686 URL: https://arxiv.org/abs/2012.09686.

[30] Sahai, P., 2020. The Communal Conflict In India Causes And Peaceful Solutions. Rajiv Gandhi Institute for Contemporary Studies, New Delhi. URL: https://www.rgics.org/wp-content/uploads/Communal-Conflict-in-India.pdf.

[31] Soruar, W., Uddin, M.M., 2021. Changes in social and religious practices of disputing communities after riot: A case study on communal violence in ramu, in: Journal of Philosophy, Culture and Religion, p. 13. doi:10.7176/JPCR/51-03.

[32] Sun, X., Gu, J., Sun, H., 2020. Research progress of zero-shot learning. Applied Intelligence 51, 36003614. doi:10.1007/s10489-020-02075-7.

[33] Tasnim, N., Gupta, S.S., Juee, F.I., Tahsin, A., Ghum, P., Fatema, K., Haque, M., Farzana, W., Nasir, P., KhudaBukhsh, A., Sadeque, F., Sushmit, A., 2024. Mapping violence: Developing an extensive framework to build a bangla sectarian expression dataset from social media interactions. URL: https://doi.org/10.48550/arXiv.2404.11752.

[34] Titli, S.R., Paul, S., 2023. Automated bengali abusive text classification: Using deep learning techniques, pp. 1–6. doi:10.1109/ICAECIS58353.2023.10170294.

[35] Windeatt, T., 2008. Ensemble MLP Classifier Design. Springer Berlin Heidelberg, Berlin, Heidelberg. pp. 133–147. doi:10.1007/978-3-540-79474-5_6.
A. Khondoker et al.: *Preprint submitted to Elsevier*  Page 13 of 13